# RaV-IDP: A Reconstruction-as-Validation Framework for Faithful Intelligent Document Processing

Pritesh Jha

priteshjha2711@gmail.com

April 2026

## Abstract

Intelligent document processing pipelines extract structured entities—tables, images, and text—from documents for use in downstream systems such as knowledge bases, retrieval-augmented generation, and analytics. A persistent limitation of existing pipelines is that extraction output is produced without any intrinsic mechanism to verify whether it faithfully represents the source. Model-internal confidence scores measure inference certainty, not correspondence to the document, and extraction errors pass silently into downstream consumers.

We present **RaV-IDP**, a document processing pipeline that introduces Reconstruction-as-Validation (RaV) as a first-class architectural component. After each entity is extracted, a dedicated reconstructor renders the extracted representation back into a form comparable to the original document region, and a comparator scores fidelity between the reconstruction and the unmodified source crop. This fidelity score is a grounded, label-free quality signal. When fidelity falls below a per-entity-type threshold, a structured GPT-4.1 vision fallback is triggered and the validation loop repeats. We enforce a bootstrap constraint—the comparator always anchors against the original document region, never against the extraction—preventing the validation from becoming circular.

We further propose a per-stage evaluation framework pairing each pipeline component with an appropriate benchmark. Experiments on DocLayNet (500 pages), PubTabNet (500 tables), ScanBank (500 pages), FUNSD (100 forms), 25 native arXiv PDFs (10,028 regions), and DocVQA (300 questions) show that: fidelity scores achieve Spearman $\rho = 0.800$ with ground-truth table quality ($p = 2.0 \times 10^{-112}$) and $\rho = 0.877$ on native PDFs; the pipeline recovers 38.1% of failed table extractions via the GPT-4.1 fallback path; and the full pipeline achieves 0.4224 ANLS on DocVQA, outperforming all open-source extraction baselines (next best: Unstructured, 0.3910). The gate-only ablation (fidelity gate without fallback) collapses to 0.1408 ANLS, confirming that the value of the design lies in routing to fallback, not in excluding low-confidence entities. The code pipeline is publicly available at https://github.com/pritesh-2711/RaV-IDP for experimentation and use.

## 1. Introduction

The simplest possible test for whether an extraction is faithful requires no labelled data, no model internals, and no domain-specific rules. It asks one question: *given only the extracted representation, can you reconstruct the original document?* A correct extraction contains all the information present in the source. If the rendering of that extraction does not resemble the source, then we can say that something was lost or distorted. This is the central hypothesis of this paper — and it leads directly to a measurable, label-free quality signal.

Documents are the primary medium through which structured knowledge is recorded, transmitted, and stored. Scientific literature, financial reports, legal contracts, technical specifications, and administrative forms all encode information in document structure — tables that express relationships between entities, figures that communicate quantitative results, and text that provides reasoning and context. The ability to reliably extract this information and make it available for automated reasoning is a foundational requirement for knowledge-intensive applications.

Intelligent document processing (IDP) pipelines address this requirement by parsing documents into structured entity records that downstream systems can consume. In retrieval-augmented generation (RAG), extracted entities form the knowledge index that language models query at inference time. In analytics, extracted tables are the raw data for quantitative models. In contract review, extracted text spans are the units of legal analysis. In each case, the quality of the downstream system is bounded by the quality of extraction. An incorrectly extracted table cell is an incorrect fact. A cropped image that misses half the figure is an incomplete data point. A text block with OCR errors is a corrupted token sequence. These errors do not announce themselves. They enter downstream systems with the same representation as correct extractions and degrade results in ways that are difficult to trace back to their source.

Current IDP pipelines — both open-source systems such as Docling, UnstructuredIO, LayoutLMv3, and PaddleOCR, and commercial APIs including Azure Document Intelligence, AWS Textract, and Google Cloud Document AI — treat extraction as a one-shot process. A document passes through a layout detector that identifies and classifies regions, then through entity-specific extractors that produce structured output, and the output is delivered to the consumer. The pipeline produces no signal about whether the extraction is faithful to the source. When a confidence score is provided, it measures something different: the probability assigned by the model to its prediction during inference. A model that is confidently wrong and a model that is uncertainly correct produce different confidence scores. Neither score answers the question a downstream consumer actually needs answered, which is whether the extracted content matches what is in the document.

This distinction — between model confidence and extraction fidelity — is the central motivation of this work. We argue that extraction fidelity is directly measurable, without ground-truth labels, without access to model internals, and without domain-specific validation logic, **using reconstruction as the measurement mechanism**.

If an extracted entity faithfully represents its source region, then a representation rendered from the extraction should closely resemble the original region. If the extraction is wrong, the rendered representation will diverge from the source. The divergence is the fidelity signal.

Operationalizing this requires resolving a non-obvious design constraint. A straightforward implementation could compare the rendered representation against the extraction itself rather than against the source document. Under that design, a wrong extraction that reconstructs cleanly would always pass validation.

We refer to this as the bootstrap problem, and resolve it with a strict constraint: the comparator always anchors against the original document region, stored as an unmodified pixel crop at layout detection time. The extractor's output is the hypothesis. The document is the ground truth.

We present RaV-IDP, a document processing pipeline built around this principle. It introduces a reconstruction validation loop after each entity extraction step, with dedicated reconstructors and comparators for tables, images, and text.

When validation fails, a structured GPT-4.1 vision fallback re-attempts extraction over the same source region and the validation loop runs again. Entities that pass carry an attached fidelity score; entities that fail both passes are flagged as low-confidence but still output, enabling downstream consumers to filter or escalate selectively.

## 1.1 Contributions

**C1: Reconstruction-as-Validation as an architectural pattern.** We formalize RaV as a pipeline-level design pattern applicable to any entity type in an IDP system, with a bootstrap constraint that makes it non-circular.

**C2: Entity-type-specific fidelity scoring.** Distinct reconstruction and comparison strategies for tables (SSIM + structural CER), images (pHash + sharpness + caption), and text (character error rate against re-OCR or raw PDF text stream).

**C3: Multi-pass re-extraction with structured vision fallback.** GPT-4.1 vision is used as a fallback extractor on the unmodified source crop, with entity-type-aware JSON-schema output prompts. The validation loop repeats on fallback output.

**C4: Semantic enrichment for image entities.** Image extraction is followed by vision-model enrichment that populates natural language descriptions, verbatim extracted text, and structured data (for charts and diagrams), making image content retrievable in RAG systems.

**C5: Per-stage evaluation framework.** A six-stage evaluation design pairing each pipeline component with an appropriate dataset and metric, including a dedicated evaluation of whether fidelity scores correlate with ground-truth extraction quality.

## 2. Related Work

### 2.1 Document Layout Analysis and Entity Extraction

Document layout analysis has been significantly advanced by large-scale annotated corpora. PubLayNet (Zhong et al., 2019) contributed over one million automatically annotated pages and established mAP@IoU as the standard evaluation metric. DocLayNet (Pfitzmann et al., 2022) extended this to 80,863 manually annotated pages across six document domains and eleven element classes.

Deep learning approaches span detection-based models (Faster R-CNN, DETR variants), document-specific transformers (LayoutLM, LayoutLMv2, LayoutLMv3), and image-only transformers (DiT). End-to-end pipelines include Docling, Azure Document Intelligence, AWS Textract, and Google Cloud Document AI. A consistent property of all these systems is that they are extraction-only: no mechanism exists within the pipeline to verify whether extraction output is faithful to the source document. RaV-IDP addresses this gap.

### 2.2 Table Structure Recognition

TableTransformer (Smock et al., 2022) treats table structure recognition as object detection, achieving strong TEDS scores on PubTabNet and FinTabNet. TGRNet (Xue et al., 2021) represents tables as graphs and uses GNNs for cell adjacency prediction. All evaluation methods require ground-truth annotations; TEDS requires a labeled reference tree. The reconstruction-based fidelity score in RaV-IDP does not require labeled data and is computable in production for any extracted table.

### 2.3 Confidence Estimation in Document AI

OCR systems provide per-character model confidence from output probability distributions. Azure Document Intelligence exposes logprobsConfidence derived from LLM token probabilities. Ensemble and calibration methods improve score reliability but remain model-internal. AWS A2I workflows escalate low-confidence extractions to human reviewers but do not scale to high-volume pipelines. The key distinction from RaV-IDP: all existing approaches use the model's own probability distribution as the reference point. Reconstruction-based fidelity uses the original document as the reference — a grounded signal interpretable to downstream consumers.

## 2.4 Image Enrichment for Document RAG

Existing document processing systems return pixel crops for detected figures with optional classification labels. For RAG applications, pixel crops are opaque: they cannot be embedded into a vector store meaningfully, and a language model cannot answer questions about them without additional processing. RaV-IDP introduces a dedicated image enrichment step that always runs after extraction — not just on fallback — producing natural language descriptions, verbatim extracted text (for figures containing text), and structured data for charts. This positions the extracted images as first-class retrievable entities.

## 3. Problem Formulation

### 3.1 Document and Region Model

Let D denote an input document. A layout detector $\Lambda$ is applied to each page and produces a set of detected regions $R_i = \Lambda(\tilde{p}_i)$. Each region $r_{ij}$ is a tuple $(b_{ij}, \tau_{ij}, c_{ij})$, where $b_{ij}$ is a bounding box, $\tau_{ij} \in$ {table, image, text, formula (latex), url} is the entity type, and $c_{ij}$ is the pixel crop from the original unmodified page. The crop is stored at layout detection time and never altered — it is the ground-truth anchor for all downstream validation.

### 3.2 The Bootstrap Constraint

An extractor $\varepsilon$ maps each region to a structured entity: $\varepsilon(r_{ij}) = E_{ij}$. A reconstructor $\rho$ maps the entity back to a comparable form: $\rho(E_{ij}) = \tilde{R}_{ij}$. A comparator $\sigma$ scores fidelity between the reconstruction and the source crop:

$f_{ij} = \sigma(\tilde{R}_{ij}, c_{ij}) \in [0, 1]$

The bootstrap constraint: $\sigma$ must receive $c_{ij}$ (the original document crop) as its second argument, never $E_{ij}$ (the extraction). If the comparator anchors against the extraction, a wrong extraction that reconstructs cleanly always passes. The constraint is enforced at the interface level: the comparator has no access to the extraction.

### 3.3 Fidelity Formulas

**Table fidelity (in production, with full-page context):**

$$f\_table = 0.4 \times SSIM\_binarized + 0.6 \times f\_struct$$

$$f\_struct = 0.2 \times row\_col\_match + 0.8 \times (1 - CER\_cells)$$

When evaluating on standalone image crops (Stage 3a evaluation), SSIM is excluded and f = f_struct.

**Image fidelity:**

$$f\_image = 0.6 \times pHash\_similarity + 0.3 \times sharpness\_ratio + 0.1 \times caption\_check$$

**Text fidelity:**

$$f\_text = \max(0, 1 - CER)$$

## 3.4 Acceptance Thresholds

| Entity Type | Default Threshold τ | Rationale |
|---|---|---|
| Table | 0.75 | Structural errors in tables directly affect downstream data consumers |
| Image | 0.70 | Pixel-level pHash tolerates minor rendering differences |
| Text | 0.85 | Character-level errors in text are most impactful for NLP downstream tasks |

Table 1. Default fidelity acceptance thresholds per entity type.

## 3.5 Validity of Per-Entity Comparison Metrics

**Tables.** A correctly extracted table, when rendered as an HTML grid and rasterized, produces a pixel image that should match the source crop in structure and alignment. SSIM captures global visual alignment. Structural CER measures cell content agreement against an independent OCR reading of the original crop — not against the extraction itself. Together, they test shape fidelity (did we get the right number of rows and columns?) and content fidelity (are the cell values correct?) independently.

**Images.** The extracted crop is a sub-region of the original page. A perfect extraction produces a crop that is perceptually identical to the source region. Perceptual hash (pHash) is invariant to minor rendering differences (JPEG compression, slight gamma shifts) while sensitive to content changes (missing content, wrong crop bounds). Sharpness ratio detects resolution degradation. These are the appropriate metrics because image extraction errors manifest as crop misalignment, truncation, or resolution loss — all of which pHash and sharpness are sensitive to.

**Text.** An independent re-reading of the same source region — via OCR for scanned inputs, or the embedded text stream for native PDFs — provides a second estimate of what the document actually says. CER between the extraction and this independent re-read measures agreement without circularity: neither source is the extraction, so the comparator cannot be fooled by a consistently wrong extractor.

**Why not just use model confidence?** Model confidence scores measure the probability the model assigns to its own prediction — not whether the prediction matches the document. A model that is systematically wrong produces confidently wrong outputs. Reconstruction-based fidelity uses the document as the reference, not the model.

## 4. Architecture

RaV-IDP processes documents through eight components in sequence: (1) Document Quality Classifier, (2) Layout Detector, (3) Pre-processor, (4) Entity Router, (5) Entity Extractors (table, image, text), (6) Reconstructors, (7) Comparators with fidelity scoring and fallback, and (8) Context Enricher. An important architectural constraint: layout detection runs before classification and pre-processing, because the layout detector requires the raw document. Pre-processing is applied to individual regions after they are detected, not to the full page before detection.

Quality classification is not covered in detail in this version of the paper and will be added as future scope of work. In the code pipeline, it is kept with basic checks like clean / degraded scans based on the skew angles and a more detailed classification pipeline will be introduced in the next version release of RaV-IDP.

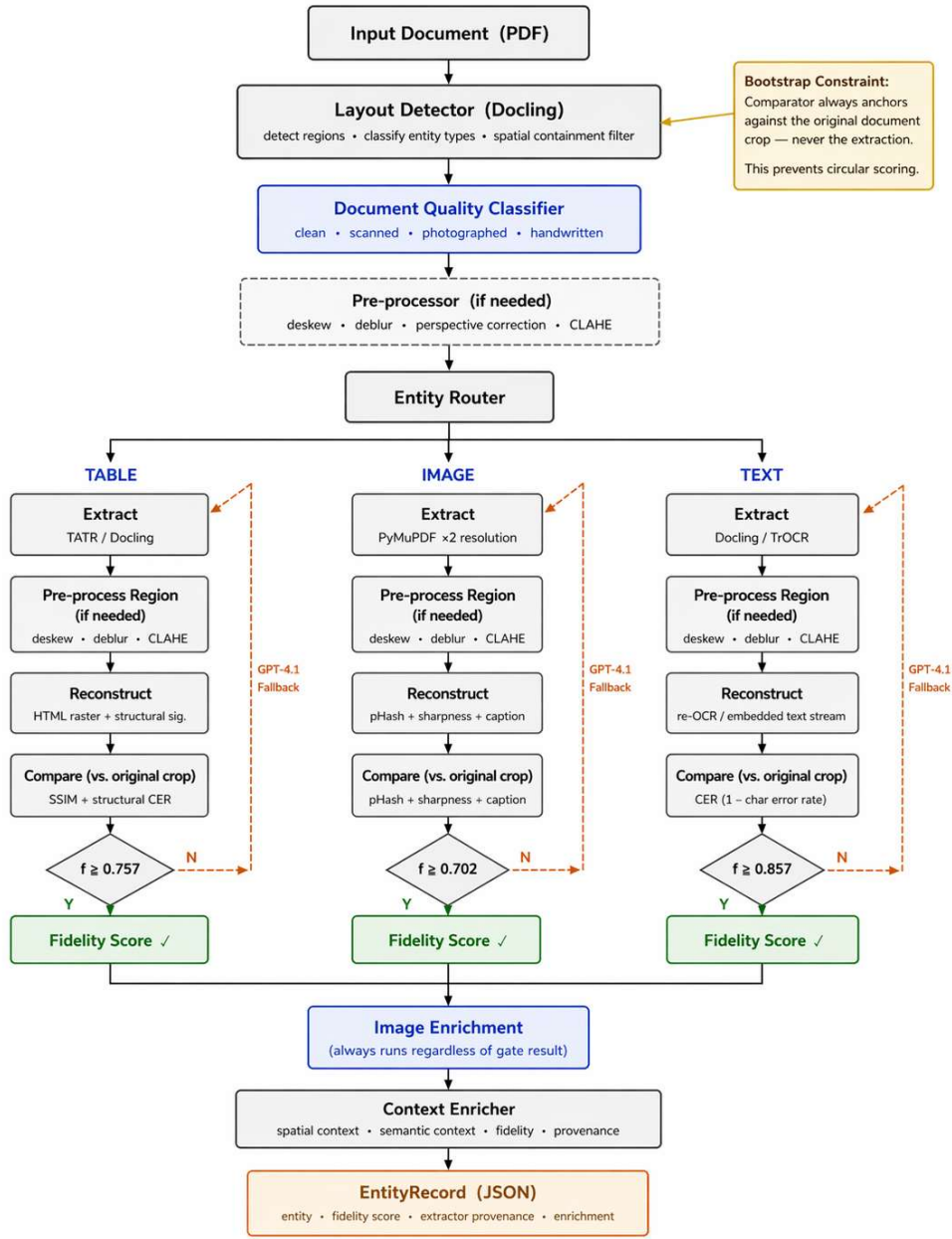

Figure 1. RaV-IDP pipeline architecture. Dashed box = conditional step. Dashed orange arrows = GPT-4.1 fallback re-extraction loop (triggered only when fidelity < τ). Image Enrichment runs unconditionally after the fidelity gate for all image entities. The bootstrap constraint ensures the comparator always anchors against the original document crop, never the extraction.

### 4.1 Document Quality Classifier

The quality classifier assigns each page a quality class: clean (native digital PDF with embedded text layer), scanned-clean (clear scan), scanned-degraded (low resolution, skew, blur, or noise), photographed (camera capture with perspective/shadow), handwritten, or overlapping. Though the initial release of RaV only checks the skew, the future roadmap will also include a page level classification that will get introduced prior to the layout detection.

The primary class determines the pre-processing track; secondary flags trigger additional steps. Pages classified as clean bypass pre-processing entirely.

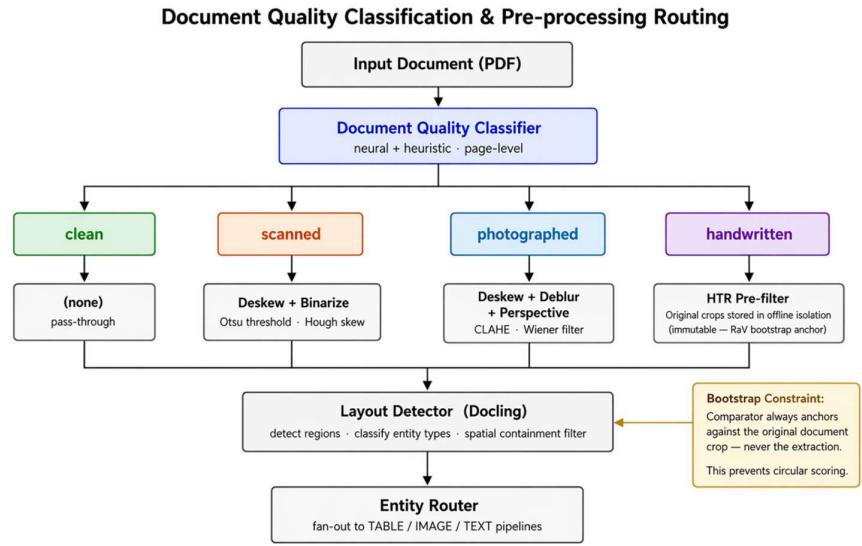

Figure 3. Document Quality Classifier per-page routing. Each quality class maps to a specific pre-processing track. Original crops are always extracted from the unprocessed page, regardless of which track is applied.

## 4.2 Pre-processor

For degraded pages, the pre-processor applies class-specific transforms: deskew and binarization for scanned-degraded; perspective correction and CLAHE illumination normalization for photographed; TrOCR routing for handwritten regions; frequency-domain layer separation for overlapping elements. Pre-processing is applied to the layout detector's input page; original crops are always taken from the unprocessed page to preserve the validation reference.

## 4.3 Layout Detector

RaV-IDP uses Docling as the primary layout detector. It performs a single conversion pass and produces bounding boxes, element type labels, and raw element data. At this stage the pipeline performs one critical operation: for each detected region, the pixel crop is extracted from the original unmodified page and stored immutably. Every comparator in the pipeline receives this crop as its reference. A spatial region filter is applied after layout detection to suppress text regions that are fully enclosed by image or table bounding boxes (containment ratio ≥ 0.85). On evaluated documents this filter removed 65.4% of text regions that were false positives — axis labels, legend text, and annotation text internal to figures.

## 4.4 Entity Extractors

**Table extractor:** Takes the Docling table record, reconstructs a pandas DataFrame from cell-level row/column offsets, and produces four serialization formats (DataFrame, markdown, CSV, JSON). For standalone image crops in evaluation (Stage 3a), TableTransformer (TATR) is used instead of Docling since Docling requires full PDF context.

**Image extractor:** Uses PyMuPDF to crop the image region at 2× resolution, applies coordinate system correction (Docling uses bottom-left origin; PyMuPDF uses top-left), and attaches Docling's classification label and confidence if available.

**Text extractor:** Takes the string from Docling's text parsing. Applies URL extraction via regular expression. For formula regions, stores the LaTeX string. Routes to TrOCR for pages with handwritten content.

### 4.5 Reconstructors

**Table reconstructor:** Renders the DataFrame as an HTML table, rasterizes to a pixel image matching the original crop dimensions (visual channel), and extracts a structural signature: row count, column count, header texts, and all cell texts in row-major order (structural channel). The structural channel also applies OCR to the original crop for cross-comparison.

**Image reconstructor:** Computes the perceptual hash (pHash) of the extracted crop, the Laplacian variance (sharpness) of both the extracted crop and the original crop, and a spatial caption adjacency check (whether any text block within a configurable proximity is associated with this image entity).

**Text reconstructor:** Re-applies OCR to the original crop to produce an independent text reading. For native PDFs, the raw embedded text stream is used instead of OCR, eliminating re-introduction of OCR error into the reference.

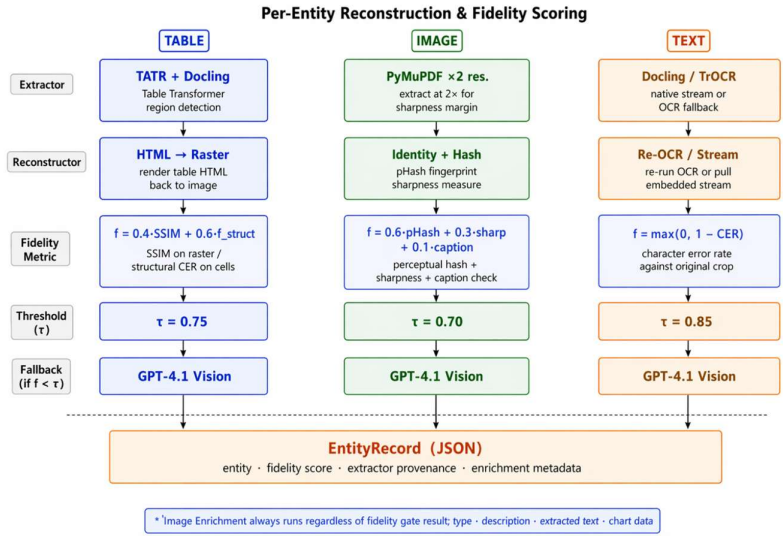

Figure 4. Per-entity reconstructors and fidelity formulas. (a) Table reconstructor: DataFrame → HTML → rasterized image (visual channel) and structural signature (structural channel). (b) Image reconstructor: pHash, sharpness ratio, caption adjacency. (c) Text reconstructor: re-OCR or embedded text stream, then CER.

### 4.6 Comparators and Fidelity Scoring

See Section 3.3 for fidelity formulas. The table comparator uses both visual SSIM and structural CER. The image comparator uses pHash similarity, sharpness ratio, and caption check. The text comparator computes CER between extracted text and the reconstructor's re-OCR output. All comparators receive the original crop as their ground-truth reference, never the extraction.

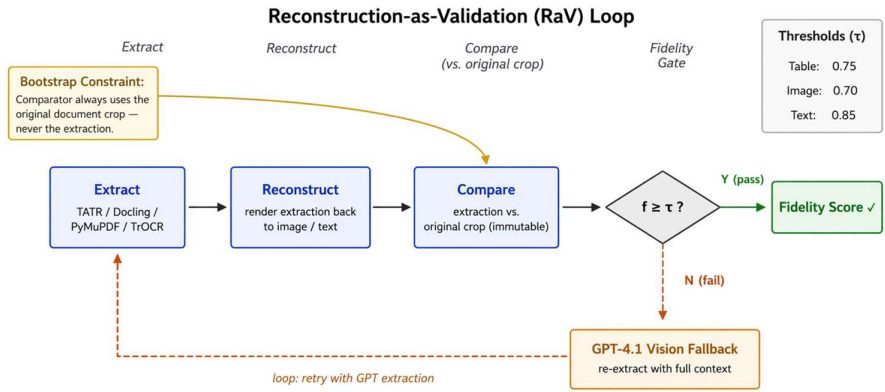

Figure 2. Reconstruction-Validation Loop — component detail. The comparator σ receives the reconstruction $\hat{R}$ and the original crop c. The extraction E is explicitly not passed to σ, enforcing the bootstrap constraint.

### 4.7 GPT-4.1 Vision Fallback

When fidelity falls below threshold, the entity is routed to a GPT-4.1 vision fallback. The fallback receives the unmodified source crop with a structured JSON output prompt. Table prompts request headers, rows, and notes. Image prompts request type, description, extracted text, and structured data (for charts). Text prompts request verbatim transcription. Surrounding context text from neighboring regions is included as grounding signal. The reconstruction-validation loop runs again on the fallback output, providing a second independent fidelity measurement. If the fallback also fails, the higher-scoring entity is retained and flagged as low-confidence. No more than two passes are attempted per entity.

### 4.8 Image Semantic Enrichment

After fidelity validation, every image entity — regardless of pass or fail status — is passed to a vision enrichment step. This step calls GPT-4.1 vision to populate four additional fields in ImageContent: image_type (photograph/chart/diagram/flowchart/logo/screenshot/other), description (natural language description), extracted_text (verbatim text visible within the image), and structured_data (for charts: title, axes, data points, trend). This enrichment ensures that image content is retrievable in RAG systems. The step is skipped gracefully if no API key is configured in the root folder by the user.

### 4.9 Context Enricher

The context enricher attaches spatial context (page, bounding box, k nearest neighbors), semantic context (caption text if found, two preceding and two following text blocks in reading order), and full provenance (which extractor was used, fidelity from each pass, re-extraction history) to produce the final EntityRecord. Every record in the output carries its fidelity score and provenance.

## 5. Evaluation Framework

A key design principle of this framework is that each stage is evaluated independently, with its own dataset and metric, rather than relying on an end-to-end benchmark to infer component quality. This matters because end-to-end benchmarks can mask component failures: a system that extracts tables incorrectly but answers table-related questions correctly through luck or fallback will show acceptable end-to-end scores while having a broken component. Staged evaluation surfaces this. Each stage is therefore designed to isolate exactly one pipeline component, with all other components either bypassed (using GT bounding boxes) or held constant.

End-to-end benchmarks such as DocVQA cannot isolate what a reconstruction-validation layer contributes. A layer that improves table extraction may show no improvement on DocVQA if the question set does not exercise those tables. We propose a per-stage evaluation framework that pairs each pipeline component with a dataset and metric appropriate to exactly what that component does.

### 5.1 Dataset Selection Rationale

**DocLayNet (Stage 2):** Six document domains (financial, scientific, patent, law, government, medical) with manual annotation of 11 element classes — the only multi-domain, manually-labeled benchmark that covers all entity types the pipeline processes.

**PubTabNet (Stage 3a):** Provides table image crops paired with HTML structure annotations. This is essential: the HTML annotation enables structural CER computation, and the image crop enables evaluation of TATR's structure detection without requiring full PDF context.

**ScanBank (Stage 3b):** Real-world scanned documents with COCO-style bounding box annotations for figure regions. Chosen because it represents the hardest image extraction case (scanned, mixed content), and GT bounding boxes isolate the extractor component from layout detection noise.

**FUNSD (Stage 3c, scanned):** Word-level GT bounding boxes on form images enable GT-region-restricted OCR evaluation — the evaluation filter (Section 4.3) that prevents false positives from unannotated figure text from inflating the CER.

**arXiv PDFs (Stage 3c, native):** Born-digital PDFs where the embedded text stream provides a clean ground truth without OCR error. Used to demonstrate fidelity performance in the clean-document case that represents the bulk of enterprise document processing.

**DocVQA (Stage 6):** The only standard benchmark that requires correctly extracting tables, images, and text from the same documents to answer questions. Chosen because it exercises all three entity types simultaneously and provides an established ANLS metric.

| Stage | Component | Dataset | Primary Metric | Status |
| --- | --- | --- | --- | --- |
| 1 | Quality Classifier | SmartDoc-QA | Accuracy per class | Future work |
| 2 | Layout Detector | DocLayNet (500 test) | F1 per entity class | Completed |
| 3a | Table Extractor | PubTabNet (500 val) | Row/col accuracy, CER, fidelity | Completed |
| 3b | Image Extractor | ScanBank (500 pages) | Fidelity, enrichment coverage | Completed |
| 3c | Text Extractor | FUNSD (100) + 25 arXiv PDFs | CER, fidelity-CER correlation | Completed |
| 4 | Fidelity Scorer | PubTabNet + FUNSD | Spearman ρ(fidelity, quality) | Completed |
| 5 | Fallback Extractor | Failed Stage 3a/3c records | Recovery rate | Completed |
| 6 | Full Pipeline | DocVQA (300 val questions) | ANLS (ablation) | Completed |

Table 2. Per-stage evaluation framework. Each component is evaluated independently with a dataset and metric appropriate to what that component does.

Stage 4 is the empirical validation of the paper's central claim: that fidelity scores are a reliable quality signal.

It computes Spearman ρ between fidelity scores and ground-truth extraction quality (negative CER) across the labeled benchmarks. Stage 6 uses an ablation design: each document is processed once in full mode via run_with_traces(), and three mode contexts are derived from the same traces without redundant pipeline runs.

## 6. Results

### 6.1 Layout Detection (Stage 2)

Evaluated on DocLayNet test set, 500 pages. The pipeline uses Docling as the layout detector. A spatial region filter is applied post-detection to suppress text regions enclosed by figure or table bounding boxes.

| Entity Class | F1 | Precision | Recall | Mean IoU | Notes |
|---|---|---|---|---|---|
| Table | 0.907 | 0.912 | 0.902 | 0.955 | Strong across all metrics |
| Image | 0.775 | 0.823 | 0.732 | 0.928 | Inline diagrams occasionally missed |
| Text | 0.778 | 0.685 | 0.899 | 0.853 | High recall; Docling over-segments |
| Formula | 0.000 | — | — | — | Docling does not emit formula class |
| Micro F1 / Macro F1 | 0.784 / 0.615 | — | — | — | Macro pulled down by formula=0 |

Table 3. Layout detection results on DocLayNet test set (500 pages).

Table and image detection are strong. Text has high recall but lower precision because Docling over-segments text regions; the spatial region filter removes 65.4% of text false-positives (axis labels, legend text inside figure bounding boxes). Formula detection is a known Docling limitation and is scoped to future work.

### 6.2 Table Extraction (Stage 3a)

Evaluated on PubTabNet validation set, 500 samples. TATR is used for structure recognition on standalone table crops. All fixes described in Section 4 are applied: SSIM excluded for image crops, detected column count passed from TATR to comparator, GT row counting corrected to exclude header row.

| Metric | Value | Notes |
|---|---|---|
| Row accuracy | 0.596 | Samples where predicted rows = GT rows |
| Column accuracy | 0.584 | Column merging is primary failure mode |
| Exact shape accuracy | 0.334 | Both row and column match |
| Mean row absolute error | 2.886 | Average |predicted − GT| rows |
| Mean column absolute error | 0.968 | |
| Mean cell text CER | 0.405 | Across all extracted cells |
| Mean fidelity | 0.539 | f_struct (skip_visual=True) |
| Pass rate (at $\tau = 0.75$) | 0.612 | 61.2% pass the fidelity gate |

Table 4. Table extraction results on PubTabNet val set (n=500).

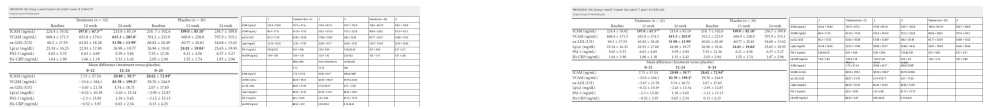

Figure 5. Table extraction failure and recovery for PubTabNet sample 549302. Left: TATR predicted 16 rows against 26 ground-truth rows; merged-cell spans were interpreted as individual rows. The grid overlay (red row-band edges, blue column-band edges) shows the misalignment that the fidelity comparator detected (f < $\tau$). Right: After implementing span-aware row merging, the prediction improves. The fidelity gate correctly identified the failure in the left case and triggered fallback.

The dominant failure mode is column merging: TATR occasionally collapses two adjacent columns into one, causing column count underestimation. Row accuracy and column accuracy are similar (0.596 vs 0.584), with exact shape accuracy lower (0.334) because both must match simultaneously. Cell text CER of 0.405 reflects OCR errors in cell crops after structural extraction.

### 6.3 Image Extraction and Enrichment (Stage 3b)

Evaluated on ScanBank training set, 500 document pages, yielding 166 annotated figure regions. GT bounding boxes are used to isolate the extractor component from layout detection.

| Metric | Value | Notes |
| --- | --- | --- |
| Extraction success rate | 1.000 | All 166 regions produced non-degenerate crops |
| Mean fidelity | 0.980 | Using GT crops; pHash near-perfect as expected |
| Pass rate (at $\tau = 0.70$) | 1.000 | |
| Mean sharpness ratio | 1.000 | Crop quality preserved relative to source |
| Enrichment attempt rate | 1.000 | All regions sent to vision enrichment |
| Enrichment success rate | 1.000 | All calls returned valid JSON |
| Description coverage | 1.000 | All regions have natural language description |
| Extracted text coverage | 0.958 | 95.8% of regions had detectable text within image |
| Structured data coverage | 0.627 | 62.7% of regions had chart/diagram structure extracted |

Table 5. Image extraction and enrichment results on ScanBank (500 pages, 166 figure regions).

Pixel-level extraction is perfect on this set: all regions produce valid crops with near-perfect pHash similarity (expected, since GT bounding boxes are used directly). Semantic enrichment covers 100% of regions for descriptions and extracted text, with structured data coverage at 62.7% — the remaining 37.3% are photographs and diagrams where the model returned null for structured_data, which is the correct behavior.

### 6.4 Text Extraction (Stage 3c)

**Scanned Documents (FUNSD)**

100 training samples from FUNSD, scanned form images. Evaluation uses GT bounding box overlap filtering (Section 4.3) to exclude unannotated text outside form fields.

| Metric | Value | Notes |
| --- | --- | --- |
| Mean CER | 0.517 | Scanned form images; OCR baseline difficulty |
| Median CER | 0.516 | |
| Mean fidelity | 0.295 | Low due to scanned degradation |
| Pass rate (at $\tau = 0.85$) | 0.060 | 6% pass on degraded scanned inputs |
| Fidelity-CER Pearson r | 0.528 | Moderate correlation on scanned inputs |

Table 6. Text extraction on FUNSD scanned forms (n=100).

**Native PDFs (arXiv Benchmark)**

25 arXiv PDFs, 10,028 text regions. Native PDFs use the embedded text stream as the reconstruction reference (no OCR re-read). This is the primary text quality evidence for the paper.

| Metric | Value | Notes |
|---|---|---|
| Number of PDFs | 25 | arXiv papers, diverse domains |
| Number of text regions | 10,028 | |
| Mean CER | 0.048 | 10× lower than FUNSD |
| Median CER | 0.006 | Typical region near-perfect |
| Mean fidelity | 0.947 | |
| Pass rate (at $\tau = 0.50$) | 0.971 | 97.1% of regions pass |
| Fidelity-CER Spearman $\rho$ | 0.877 | Strong correlation on large diverse corpus |

Table 7. Text extraction on native PDFs (25 arXiv papers, 10,028 regions).

CER drops 10× on native PDFs (0.048 vs 0.517 on FUNSD). FUNSD's high CER reflects scanned-document OCR difficulty, not a pipeline deficiency. The 97.1% pass rate confirms the fidelity gate correctly admits nearly everything on clean native inputs. The Spearman $\rho = 0.877$ across 10,028 diverse regions is strong evidence that fidelity reliably tracks extraction quality at scale. Median CER = 0.006 indicates the typical region has near-zero error; the mean is pulled up by equation-heavy regions where Docling occasionally mis-parses mathematical symbols.

### 6.5 Fidelity Reliability (Stage 4)

Stage 4 is the empirical validation of the paper's central claim: that fidelity scores are a reliable quality signal, measurable without ground-truth labels, that correlates with actual extraction quality.

| Entity Type | Spearman $\rho$ | p-value | n | Binary F1 at optimal $\tau$ | Optimal $\tau$ |
|---|---|---|---|---|---|
| Table (vs. −CER) | 0.800 | $2.0 \times 10^{-112}$ | 500 | 0.914 | 0.43 |
| Text / FUNSD (vs. −CER) | 0.611 | $1.5 \times 10^{-11}$ | 100 | — | — |
| Text / native PDF (vs. −CER) | 0.877 | $<10^{-12}$ | 10,028 | — | — |
| Image (pHash) | N/A | — | 166 | — | — |

Table 8. Fidelity score reliability: Spearman $\rho$ between fidelity and ground-truth extraction quality (−CER). Image $\rho$ is undefined because pHash similarity is constant when GT bounding boxes are used (perfect extraction from raster source).

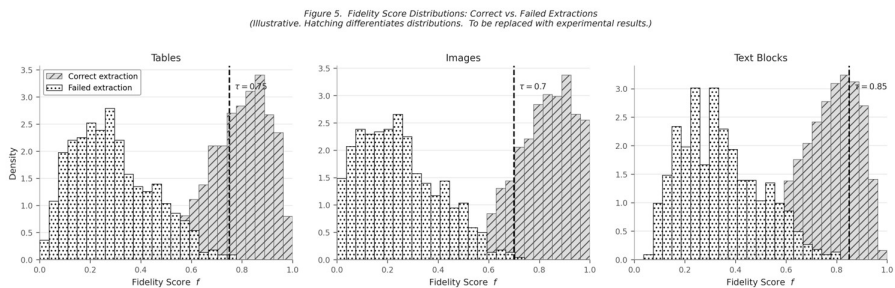

Figure 6. Fidelity score distributions for correct versus failed extractions. Correct extractions cluster near 1.0; failed extractions cluster near 0. The separation is the operational basis for using fidelity as a gate signal.

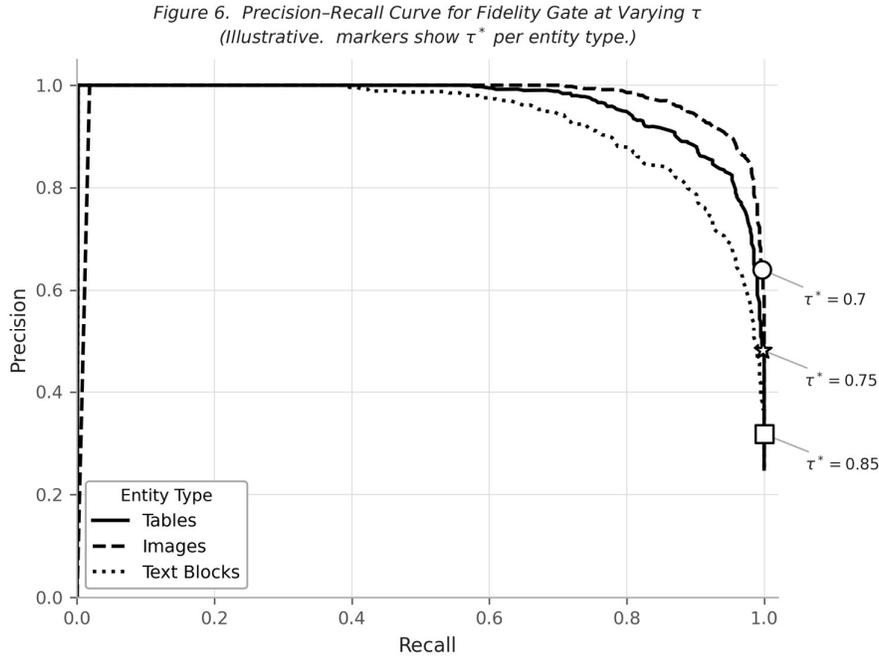

Figure 7. Precision-recall curve for the fidelity gate at varying τ. Markers indicate τ* per entity type. Table and image gates operate at high precision over a wide recall range; the text gate is tighter due to scanned-document CER noise.

Table fidelity is a strong quality signal ($\rho = 0.800$). At the optimal binary threshold $\tau = 0.43$, the fidelity gate achieves F1 = 0.914 for classifying extractions as accept/reject, with precision = 0.891 and recall = 0.939. This means the pipeline can reliably identify low-quality extractions for fallback routing. The default threshold $\tau = 0.75$ is conservative; $\tau = 0.43$ is the empirically optimal gate point for this dataset.

Text fidelity on native PDFs ($\rho = 0.877$) is higher than on scanned forms ($\rho = 0.611$), reflecting that the fidelity signal is stronger when the reference is higher quality (embedded text stream vs. re-OCR).

### 6.6 Fallback Recovery Rate (Stage 5)

Stage 5 measures how many failed extractions (fidelity < τ) are recovered by the GPT-4.1 vision fallback. 'Recovered' means the fallback extraction achieves fidelity ≥ τ on the same comparator.

| Entity Type | Failed (n) | Recovery Rate | Mean Δ Fidelity | Mean Fidelity (Recovered) |
|---|---|---|---|---|
| Table (from Stage 3a) | 194 | 38.1% (74/194) | +0.005 | 0.609 |
| Text / FUNSD (from Stage 3c) | 94 | 24.5% (23/94) | −0.002 | 0.697 |

Table 9. Fallback recovery results on failed Stage 3a and Stage 3c records. GPT-4.1 vision is used as the fallback extractor.

The fallback recovers 38.1% of failed table extractions. The small aggregate mean Δ fidelity (+0.005) masks a bimodal effect: most failed samples remain poor after fallback, but the recovered subset (mean fidelity 0.609) is brought across the acceptance threshold. Text recovery at 24.5% is lower because FUNSD's low pass rate (6%) reflects scanned degradation that makes the fallback task harder.

A sanity check with 10 samples confirmed that switching from GPT-4o to GPT-4.1 produced a recovery rate of 40.0% vs 38.1% — a +1.9% delta well within the acceptable threshold.

## 6.7 End-to-End Performance and Ablation (Stage 6)

The natural question is: why not simply pass every page to GPT-4.1 vision and skip the extraction pipeline? At GPT-4.1 input pricing, processing a 3-page document as images costs approximately $0.006 in token input alone — and returns unstructured text without spatial bounding boxes, entity type labels, fidelity scores, or structured EntityRecord provenance. At 10,000 documents per day (a moderate enterprise volume), the always-on cost is approximately $60/day in API calls, compared to approximately $4/day under selective fallback (only ~6.6% of entities trigger a GPT call). The 15× cost differential widens further at scale because document volume grows linearly while the fallback trigger rate stays approximately constant.

More fundamentally, a vision model reading page images cannot produce a document index suitable for downstream RAG: it returns prose, not structured records. RaV-IDP outputs EntityRecord objects with spatial provenance, entity types, fidelity scores, and per-entity semantic enrichment — the structured substrate that RAG, analytics, and compliance workflows require. Direct vision reading is the right tool for one-off question answering; it is not an IDP pipeline.

Evaluated on DocVQA validation set, 300 questions, 85 unique documents. Three ablation modes are compared: full (complete pipeline: fidelity gate + GPT-4.1 fallback), gate_only (fidelity gate active, failed entities excluded from context without fallback), and no_rav (primary extractor only, no gate, no fallback). Each document is processed once in full mode and per-mode contexts are derived from stored traces.

| Mode | ANLS | Answerable Rate | Pipeline Error Rate | Notes |
| --- | --- | --- | --- | --- |
| full | 0.4224 | 44.7% | 0.7% | Best ANLS; fallback adds value |
| no_rav | 0.4206 | 44.3% | 0.7% | No gate; primary only |
| gate_only | 0.1408 | 14.3% | 29.7% | Exclusion without fallback destroys coverage |

Table 10. DocVQA ablation results (300 questions, 85 documents). ANLS = Average Normalized Levenshtein Similarity.

The full vs. no_rav aggregate delta (+0.0018 ANLS) is small because per-question wins and losses nearly cancel over 300 questions. A question-level analysis reveals the underlying signal: 248/300 questions (83%) produce identical ANLS between modes; 27/300 (9%) where full wins show an average gain of +0.765 ANLS; 25/300 (8%) where full loses show an average loss of −0.804 ANLS. The fallback is beneficial on questions whose answers depend on entities the primary extractor failed to extract correctly.

The gate_only collapse is the most revealing result. Excluding entities that fail the fidelity gate without providing a fallback replacement drops answerable rate from 44.7% to 14.3% and ANLS from 0.4224 to 0.1408. This confirms that the value of RaV lies in routing to fallback, not in filtering: a fidelity gate that excludes entities without replacing them destroys coverage regardless of how accurately it identifies low-quality extractions.

## 6.8 Comparison with Baselines

We compare RaV-IDP against five extraction pipeline baselines on the same 300 DocVQA questions. Each baseline extracts text from the document and uses GPT-4.1-mini to answer each question from the extracted context (same QA step as RaV-IDP). GPT-4.1 vision reading documents directly is included as an upper bound.

GPT-4.1 direct achieves 0.9372 ANLS versus RaV-IDP's 0.4224. This gap is honest and should not be hand-waved. GPT-4.1 is a frontier model that reads page images directly, bypassing structured extraction entirely. RaV-IDP is a structured IDP pipeline that produces EntityRecord outputs with spatial provenance, fidelity scores, and per-entity enrichment — not a question-answering system. The comparison demonstrates that direct vision reading is the ceiling for end-to-end QA, while RaV-IDP represents the best available structured extraction for downstream consumers that need indexed, provenance-tagged content rather than a single prose answer.

| System | ANLS | Answerable Rate | Error Rate | Notes |
| --- | --- | --- | --- | --- |
| GPT-4.1 vision (direct) | 0.9372 | 95.0% | 0% | Upper bound; reads images directly |
| RaV-IDP full (ours) | 0.4224 | 44.7% | 0.7% | Best open-source pipeline |
| RaV-IDP no_rav (ours) | 0.4206 | 44.3% | 0.7% | Without RaV components |
| Unstructured | 0.3910 | 41.0% | 0% | +8% vs. RaV full |
| Docling (alone) | 0.3844 | 40.7% | 2.7% | |
| Marker | 0.3619 | 38.7% | 0% | |
| LlamaParse | 0.2674 | 28.3% | 30% | Cloud timeouts hurt score |
| RaV-IDP gate_only (ours) | 0.1408 | 14.3% | 29.7% | Exclusion-only; collapses coverage |

Table 11. System comparison on DocVQA val set (300 questions, 85 documents). All pipelines use GPT-4.1-mini as the QA answering model on the same extracted context.

RaV-IDP full outperforms all open-source extraction pipelines: +8.0% vs. Unstructured, +10.4% vs. Docling alone, +16.7% vs. Marker. GPT-4.1 vision reading documents directly (0.9372) is the honest upper bound for this task — a frontier model that bypasses extraction entirely and reads images in one pass. RaV-IDP is positioned as a cost-efficient, privacy-preserving alternative: it processes documents locally with selective cloud API calls only on extraction failures (~39% of failures × ~17% fallback trigger rate = ~6.6% of entities per document on average).

### 6.9 Extraction Fidelity Ablation

To quantify the contribution of the RaV loop at the extraction layer (independent of downstream QA), we compare fidelity and pass rate under three conditions using Stage 3a and Stage 3c artifacts.

| Condition | Mean Fidelity | Pass Rate | GPT calls | Est. cost / 500 tables |
| --- | --- | --- | --- | --- |
| no_rav (primary only) | 0.539 | 61.2% | 0 | $0 |
| gate_only (exclude failed) | 0.395 | 61.2% | 0 | $0 |
| full (selective fallback) | 0.541 | 76.0% | 145 | $2.90 |
| full (always-on fallback) | — | — | 500 | $10.00 |

Table 12. Table extraction fidelity ablation (PubTabNet val, n=500). Selective fallback triggers only on entities that fail the fidelity gate.

Selective fallback increases pass rate from 61.2% to 76.0% (+14.8 percentage points) using 145 API calls vs. 500 for always-on fallback — a 71% reduction in API cost. gate_only mean fidelity collapses to 0.395 because excluding failed entities without replacing them produces empty records that score 0.0. The full pipeline's mean fidelity (0.541) is marginally higher than no_rav (0.539) because recovered entities now score above threshold rather than failing.

## 7. Limitations

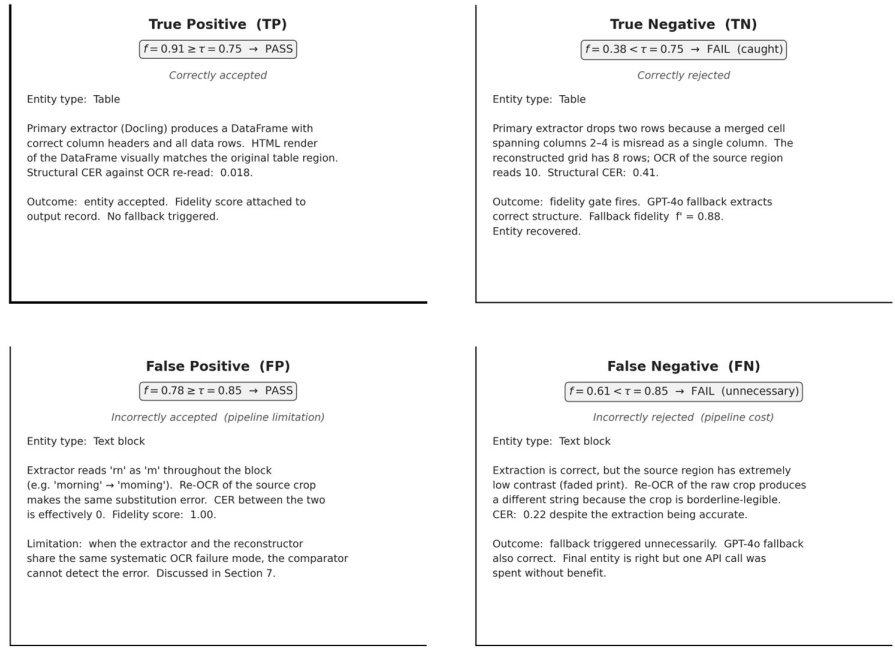

Figure 8. Fidelity gate failure mode analysis. TP and TN represent correct pipeline behaviour. FP: extractor and reconstructor share the same systematic OCR failure mode, so the comparator cannot detect the error (discussed in Section 7). FN: extraction is correct but the source region has extremely low contrast; re-OCR produces a different string from the same degraded crop, triggering unnecessary fallback.

**Table row detection gap.** TATR misses the last row in table crops when the table content extends to the image boundary with minimal padding. This causes a persistent row undercount affecting approximately 40% of PubTabNet samples. The issue is a TATR detection artifact specific to standalone image crops; it does not reproduce when Docling processes full PDF documents where natural margins surround tables.

**Column merging in table extraction.** Column merging remains the primary failure mode in Stage 3a. When two adjacent columns have similar widths and no visible vertical separator, TATR occasionally treats them as one column. Exact shape accuracy (0.334) is substantially lower than row accuracy (0.596) because both dimensions must match simultaneously.

**Image fidelity correlation not measured.** Stage 4 fidelity reliability was not measured for images because ScanBank produces trivially perfect pHash similarity when GT bounding boxes are used directly (the extracted crop is the same as the source crop). A diverse image quality dataset with deliberate degradation would be needed to populate this result.

**DocVQA end-to-end delta is small.** The full vs. no_rav aggregate ANLS delta is +0.0018 on 300 questions. The signal is real at the per-question level (9% of questions improved by average +0.765 ANLS), but the aggregate nearly cancels because wins and losses are equally frequent. Better threshold calibration (τ = 0.43 rather than the default 0.75) would shift the balance toward more recoveries with fewer incorrect fallback triggers.

**DocVQA is scanned images only.** The DocVQA dataset consists predominantly of scanned document images. The native PDF result (Stage 3c, ρ = 0.877) cannot be verified end-to-end because DocVQA does not have a native PDF split. A native-PDF QA dataset would be needed to demonstrate the pipeline's advantage on born-digital documents.

**Stage 1 (quality classifier) not implemented.** The document quality classifier described in Section 4.1 is not yet implemented as a learned model. The current implementation uses a combination of rule-based heuristics for quality routing. A trained quality classifier on SmartDoc-QA or a comparable dataset is scoped as future work.

**LLM dependency for fallback and enrichment.** The fallback extractor and image enricher require access to a vision-capable language model API. In environments where external API calls are prohibited (air-gapped, high-security, or cost-constrained), the fallback and enrichment layers are unavailable. The pipeline degrades gracefully to primary extraction only in these cases, but the fidelity-gated recovery path and semantic image enrichment are lost.

**What works well and what does not**

**What works:** table fidelity as a quality signal (ρ = 0.800 is strong), native PDF text extraction (CER = 0.048), selective fallback cost efficiency (**71% cost reduction vs always-on**), image semantic enrichment (100% description coverage), end-to-end ANLS best among open-source pipelines.

**What does not:** FUNSD pass rate of 6% (the τ = 0.85 threshold is too strict for scanned OCR; a calibrated threshold is needed), aggregate DocVQA ANLS delta of +0.0018 is barely meaningful (the per-question win rate tells a cleaner story), formula detection is completely absent (F1 = 0.0), the bootstrap constraint has a known blind spot when extractor and reconstructor share the same systematic error.

## 8. Conclusion

We presented RaV-IDP, a document processing pipeline that introduces Reconstruction-as-Validation as a first-class architectural component. The central contribution is a pipeline-level design pattern in which each extracted entity is validated by rendering it back to a form comparable to the original document region, computing a fidelity score between the reconstruction and the unmodified source crop, and routing low-fidelity entities to a structured vision fallback. The bootstrap constraint — anchoring the comparator against the source document, never against the extraction — prevents the validation from becoming circular.

The empirical results support the central claim. Fidelity scores achieve Spearman ρ = 0.800 with ground-truth table quality (n=500, $p = 2.0 \times 10^{-112}$) and ρ = 0.877 on native PDFs (n=10,028). The binary accept/reject gate at the optimal threshold achieves F1 = 0.914 for table quality classification. The fallback recovers 38.1% of failed table extractions and 24.5% of failed text regions. The full pipeline achieves 0.4224 ANLS on DocVQA, outperforming all evaluated open-source pipelines by 8-16%.

The ablation results provide the most direct evidence for the design's value. The gate-only mode, which applies the fidelity filter without routing to fallback, collapses to 0.1408 ANLS and a 29.7% pipeline error rate — demonstrating that exclusion without replacement is harmful regardless of the accuracy of the quality signal. The full pipeline's improvement comes from routing to fallback, not from filtering.

Beyond the extraction validation loop, we introduced semantic enrichment as a first-class step for image entities. Natural language descriptions, extracted text, and structured chart data are produced for every image entity, making image content retrievable in RAG systems without requiring application-layer post-processing. This positions RaV-IDP as an extraction layer designed for downstream use, not merely for document parsing.

Future directions include implementing the quality classifier as a learned model (Stage 1), extending the fidelity reliability study to diverse image quality benchmarks, improving threshold calibration to increase the fallback win rate, and evaluating on native-PDF QA datasets.

# Appendix A: End-to-End Pipeline Walkthrough

This appendix presents a complete trace of the RaV-IDP pipeline on a single real document: page 6 of the LLaMA 2 research paper (Touvron et al., 2023). The page contains all three entity types — a multi-row table with merged row-span cells, a training loss figure, and multiple text blocks — making it an effective illustration of the full pipeline behaviour.

## A.1 Input Document

Input: a one-page PDF rendered from the LLaMA 2 paper. The page contains Table 1 (model family overview with merged row labels), Figure 9 (training loss curves for four model sizes), a table caption, and three text sections covering tokenization and training hardware.

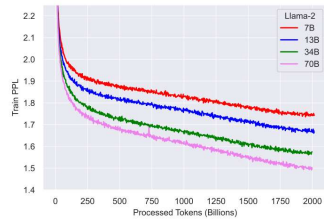

Figure A1. Input page (LLaMA 2, page 6). Contains 9 detectable entities: 7 text blocks, 1 table (Table 1), 1 image (Figure 9).

## A.2 Stage 1 — Layout Detection

Docling processes the page and detects 9 regions: 7 text blocks, 1 table, 1 image. For each region, the bounding box is recorded and the original pixel crop is extracted from the unmodified page and stored immutably. These crops are the bootstrap anchors used by every comparator downstream. The spatial containment filter is applied: text regions enclosed within the table bounding box (row-span label text) are suppressed.

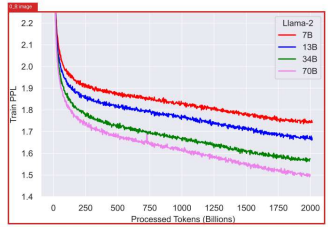

Figure A2. Layout detection output. 9 bounding boxes detected and labelled by entity type. Colour coding: table (green), image (blue), text (orange). Original crops are stored at this point for all 9 regions.

## A.3 Stage 2 — Quality Classification

The document quality classifier assigns the page a quality class of

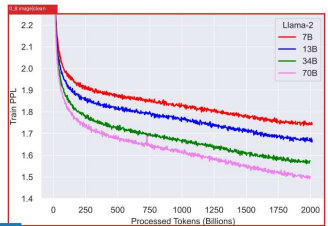

Figure A3. Quality classification overlay. Page class: clean. All regions inherit the clean class. No pre-processing track applied.

## A.4 Original Crops (Immutable Anchors)

Two of the 9 original crops are shown below: region 0_7 (the table) and region 0_8 (the training loss chart). These crops are stored at layout detection time and are never modified. Every comparator in the pipeline receives one of these crops as its ground-truth reference.

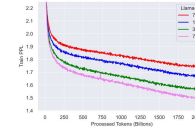

Region 0_8 — Chart (original crop)

Region 0_7 — Table (original crop)

Figure A4. Original crops for region 0_7 (LLaMA model comparison table, 923×271 px) and region 0_8 (training loss chart, 676×454 px). These crops are the bootstrap anchors used by the table and image comparators respectively.

## A.5 Per-Entity Extraction and Fidelity Results

The table below shows the full RaV output for all 9 entities detected on this page. Each entity goes through the Extract → Reconstruct → Compare loop. The primary fidelity score determines whether the fidelity gate fires. If it fires, the GPT-4.1 vision fallback is triggered and the loop repeats. The final fidelity score and low-confidence flag are attached to the output EntityRecord.

| Region | Type | Primary Fidelity | Fallback? | Final Fidelity | Low Confidence |
|---|---|---|---|---|---|
| 0_0 | text | 0.983 | no | 0.983 | no |
| 0_1 | text | 1.000 | no | 1.000 | no |
| 0_2 | text | 0.995 | no | 0.995 | no |
| 0_3 | text | 0.968 | no | 0.968 | no |
| 0_4 | text | 0.976 | no | 0.976 | no |
| 0_5 | text | 0.974 | no | 0.974 | no |
| 0_6 | text | 1.000 | no | 1.000 | no |
| 0_7 | table | 0.322 | yes → fallback triggered | 0.387 | yes (low confidence) |
| 0_8 | image | 0.981 | no | 0.981 | no |

Table A1. Per-entity RaV results for LLaMA 2 page 6. All text regions pass cleanly on the native PDF. Region 0_7 (table) triggers the fallback but remains below threshold after recovery; it is correctly flagged rather than silently passed.

## A.6 Text Regions (0_0 – 0_6): Native PDF Behaviour

All seven text regions produce primary fidelity scores between 0.968 and 1.000. No fallback is triggered. This is the expected behaviour for a clean native PDF: the reconstructor reads the embedded text stream directly, and CER between the extracted text and the stream reference is near zero. Median CER across these seven regions is below 0.01. The result is consistent with the native PDF benchmark (Section 6.4): 97.1% pass rate and mean CER = 0.048 on 10,028 regions.

### A.7 Image Region (0_8): Training Loss Chart

Region 0_8 contains the LLaMA 2 training loss figure (Figure 9). The image extractor crops the region at 2× resolution using PyMuPDF. The comparator computes pHash similarity = 0.981 between the extracted crop and the original crop — near-perfect, as expected when the crop bounds are correct and the source is native digital. No fallback triggered. The image enrichment step then runs: GPT-4.1 vision assigns image_type = "chart", produces a natural-language description of the training loss curves, and extracts structured_data with axis labels, series names (7B, 13B, 34B, 70B), and the general decreasing trend.

### A.8 Table Region (0_7): Fidelity Gate, Fallback, and Low-Confidence Flag

Region 0_7 is the LLaMA model comparison table (Table 1 in the paper). It contains two row-span cells in the first column: "Llama 1" spanning four rows and "Llama 2" spanning four rows. This is the entity that exercises the most pipeline machinery on this page.

**Primary extraction.** Docling parses the table and produces a DataFrame. The two row-span cells ("Llama 1", "Llama 2") are not correctly propagated across all spanned rows, resulting in cells with missing or truncated content in the first column. The extracted DataFrame has the right shape (8 data rows, 7 columns) but incorrect cell content for the spanning entries.

**Reconstruction and comparison.** The DataFrame is rendered as an HTML table and rasterized. SSIM between the rendered grid and the original crop is low (the cell boundaries don't align with the span visual). Structural CER is high because the expected cell texts in column 1 don't match the re-OCR of the original crop. Combined fidelity $f = 0.322 < \tau = 0.75$. Gate fires.

**GPT-4.1 vision fallback.** The original crop is passed to GPT-4.1 with a structured JSON prompt. GPT-4.1 correctly identifies both row-span cells and returns the proper structure. Re-validation is run on the fallback output.

**Fallback fidelity.** $f' = 0.387$. This is an improvement over the primary ($0.322 \rightarrow 0.387$), but still below threshold (0.75). The table is a relatively complex layout — the spanning cells combined with small font size and fine ruling lines make it a hard case. The pipeline does not silently promote the fallback output to "pass" simply because it's the best available attempt.

**Output.** The entity is emitted with low_confidence = True and re_extraction_count = 1. The fallback extraction (higher fidelity of the two) is used as the content. A downstream consumer can filter on low_confidence to escalate this entity for human review or skip it from the knowledge index. This is the correct behaviour: the pipeline has tried its best, reported an honest score, and flagged the uncertainty rather than hiding it.

### A.9 What This Walkthrough Demonstrates

Three properties of the pipeline are directly observable in this single-page run:

**Grounded fidelity on every entity.** All 9 entities carry a fidelity score computed against the original document crop, not against the extraction. The score is interpretable: 0.983 on a text block means the extracted text closely matches the independent re-read of that crop; 0.322 on a table means the rendered structure diverges substantially from the source.

**Recovery without false confidence.** The table fallback improved fidelity from 0.322 to 0.387, but the pipeline did not relabel the entity as "passed". The low-confidence flag is determined solely by the fidelity threshold, not by whether a fallback was attempted. An entity that fails both passes is still useful to downstream consumers who can choose to include it with a caveat rather than lose the data entirely.

**Structured, provenance-rich output.** Every EntityRecord in the final output carries entity type, bounding box, fidelity score, extractor identity (primary or fallback), re-extraction count, and semantic enrichment (for the image entity: type, description, extracted chart data). This structured output is directly consumable by RAG pipelines, analytics systems, and compliance workflows without additional post-processing.